\documentclass[sigconf]{acmart}
\AtBeginDocument{%
  }

\copyrightyear{2025}
\acmYear{2025}
\setcopyright{rightsretained}\acmConference[COLIEE 2025]{COLIEE 2025 workshop: Competition on Legal Information Extraction/Entailment}{June 20, 2025}{Chicago, USA}
\acmBooktitle{Proceedings of COLIEE 2025 workshop, June 20, 2025, Chicago, USA}
\acmPrice{}
\acmISBN{}
\acmDOI{}

\usepackage[normalem]{ulem}
\usepackage{listings}
\usepackage{xcolor} 
\usepackage{algorithm}
\usepackage{pifont}
\usepackage{algpseudocode}
\useunder{\uline}{\ul}{}
\usepackage{subcaption}
\usepackage{array}

\begin{document}

\title[NOWJ@COLIEE 2025]{NOWJ@COLIEE 2025: A Multi-stage Framework Integrating Embedding Models and Large Language Models for Legal Retrieval and Entailment}

%
\author{Hoang-Trung Nguyen}
\orcid{0009-0009-0758-441X}
\affiliation{%
  \institution{VNU University of Engineering and Technology}
  \city{Hanoi}
  \country{Vietnam}
}
\email{20020083@vnu.edu.vn}

\author{Tan-Minh Nguyen}
\orcid{}
\affiliation{%
  \institution{Japan Advanced Institute of Science and Technology}
  \city{Ishikawa}
  \country{Japan}
}
\email{minhnt@jaist.ac.jp}

\author{Xuan-Bach Le}
\orcid{0009-0000-1527-7174}
\affiliation{%
  \institution{VNU University of Engineering and Technology}
  \city{Hanoi}
  \country{Vietnam}
}
\email{22024506@vnu.edu.vn}

\author{Tuan-Kiet Le}
\orcid{}
\affiliation{%
  \institution{VNU University of Engineering and Technology}
  \city{Hanoi}
  \country{Vietnam}
}
\email{22024546@vnu.edu.vn}

\author{Khanh-Huyen Nguyen}
\orcid{}
\affiliation{%
  \institution{VNU University of Engineering and Technology}
  \city{Hanoi}
  \country{Vietnam}
}
\email{22026502@vnu.edu.vn}

\author{Ha-Thanh Nguyen}
\orcid{0000-0003-2794-7010}
\affiliation{%
  \institution{Center for Juris-Informatics, ROIS-DS}
  \city{}
  \country{}
}
\affiliation{%
  \institution{Research and Development Center for Large Language Models, NII}
  \city{Tokyo}
  \country{Japan}
}
\email{nguyenhathanh@nii.ac.jp}

\author{Thi-Hai-Yen Vuong}
\orcid{0000-0002-8019-7178}
\affiliation{%
  \institution{VNU University of Engineering and Technology}
  \city{Hanoi}
  \country{Vietnam}
}
\email{yenvth@vnu.edu.vn}

\author{Le-Minh Nguyen}
\orcid{0000-0002-2265-1010}
\affiliation{%
  \institution{Japan Advanced Institute of Science and Technology}
  \city{Ishikawa}
  \country{Japan}
}
\email{nguyenml@jaist.ac.jp}







\renewcommand{\shortauthors}{Hoang-Trung Nguyen et al.}

\begin{abstract}
This paper presents the methodologies and results of the NOWJ team's participation across all five tasks at the COLIEE 2025 competition, emphasizing advancements in the Legal Case Entailment task (Task 2). Our comprehensive approach systematically integrates pre-ranking models (BM25, BERT, monoT5), embedding-based semantic representations (BGE-m3, LLM2Vec), and advanced Large Language Models (Qwen-2, QwQ-32B, DeepSeek-V3) for summarization, relevance scoring, and contextual re-ranking. Specifically, in Task 2, our two-stage retrieval system combined lexical-semantic filtering with contextualized LLM analysis, achieving first place with an F1 score of 0.3195. Additionally, in other tasks--including Legal Case Retrieval, Statute Law Retrieval, Legal Textual Entailment, and Legal Judgment Prediction--we demonstrated robust performance through carefully engineered ensembles and effective prompt-based reasoning strategies. Our findings highlight the potential of hybrid models integrating traditional IR techniques with contemporary generative models, providing a valuable reference for future advancements in legal information processing.
\end{abstract}



\received{7 April 2025}
\received[accepted]{5 May 2025}

\maketitle

\section{Introduction}
Legal text processing is a specialized field that requires knowledge of both law and information science. The use of artificial intelligence (AI) and large language models (LLMs) as supporting tools in judicial processes has become more prevalent \cite{zhong-etal-2020-nlp,MORENOSCHNEIDER2022101966,10.1145/3627673.3680020}. COLIEE \cite{coliee_2024} is an annual event organized to support the research of legal information processing. The competition covers various challenges, including document retrieval, textual entailment, and judgment prediction. COLIEE is a valuable opportunity for researchers to explore and evaluate various advanced techniques in complex real-world judicial problems. 

The COLIEE 2025 competition comprises five tasks that span two major legal systems: case law and statute law. Tasks 1, 2, and 5 focus on case law, drawing on legal cases from the Federal Court of Canada and Japanese court decisions. Task 1 (Legal Case Retrieval) involves identifying relevant precedents that support the decision of a given case, serving as a foundational step for Task 2 (Legal Case Entailment), which aims to determine whether specific paragraphs within the retrieved cases entail the decision. Task 5 (Legal Judgment Prediction) targets the prediction of judicial outcomes in real-world civil cases from Japanese courts. In contrast, Tasks 3 and 4 are based on the statute law system, which is followed by countries such as Japan, Vietnam, and many European nations. Task 3 (Statute Law Retrieval) requires participants to retrieve statutory articles from the Japanese Civil Code corpus that support a given legal query. Building upon this, Task 4 (Legal Textual Entailment) challenges systems to provide a binary answer—``Yes'' or ``No''—based on the retrieved legal content.

COLIEE 2025 is the third year the NOWJ team has participated. We propose multi-state frameworks based on state-of-the-art models such as monoT5, Llama-3, Qwen-2, DeepSeek, and prompting techniques for handling complex legal tasks of the competition. Specifically, a four-state framework involving meticulous pre-processing, LLM-based summarization, and different retrieval approaches is proposed for Task 1. For Task 2, we combined the advantages of lexical matching, semantic ranking methods, and recent LLMs (e.g., QwQ, DeepSeek) to model the entailment relationships between judicial factors. For Task 3, we leveraged pre-trained language models to develop a retrieval method based on bi-encoder and cross-encoder models. Various LLMs and prompting techniques, including zero-shot and few-shot prompts, are explored in Task 4. Finally, we employed a hierarchical language model, clustering approach, and heuristic post-processing phase for addressing real-world judgment prediction problems in Task 5. Our proposed methods achieved promising results, including first place in Task 2, and a third rank on the Tasks 3 and 5 leaderboard. Our findings highlight the potential of hybrid models integrating traditional IR techniques with contemporary generative models, providing a valuable reference for future advancements in legal information processing.

The remainder of this paper presents our methods used in the competition, with each section dedicated to a specific task. The final section concludes the paper and outlines directions for future research. The source code will be released shortly to support the reproducibility of our work.

 
\section{Task 1: Legal Case Retrieval}

\subsection{Task Overview}
Many countries, including Canada and the United States, follow the common law system, where case law is a fundamental component of judicial practice. Judges and legal professionals rely on precedents when handling new cases. To advance research in legal processing, the Legal Case Retrieval task aims to identify noticed cases—previous cases that support a given case decision. Specifically, this task requires extracting all supporting cases $\{d_1, d_2,\dots, d_n\}$ that are semantically or logically similar to a given query case $q$.
A standard case document consists of four main sections: background, facts, reasoning, and decision. However, the lack of a unified document structure makes case presentation and extraction challenging. Additionally, case documents typically range from 4,000 to 10,000 words, exceeding the input limits of many pre-trained language models (e.g., BERT, T5), making efficient processing difficult.



Last year, the winning team, TQM \cite{10.1007/978-981-97-3076-6_15}, combined lexical and dense models to generate features and improve case relevance understanding. They also applied meticulous pre-processing and post-processing to filter out irrelevant information.
The runner-up team, UMNLP \cite{10.1007/978-981-97-3076-6_11}, proposed a pairwise similarity ranking framework at multiple levels, including paragraphs, sentences, and ``propositions''. They trained a multilayer perceptron model to assess case relevance using various features extracted from each query-candidate pair.

\subsection{Methodology}
\begin{figure}[t]
    \centering
    \includegraphics[width=1\linewidth]{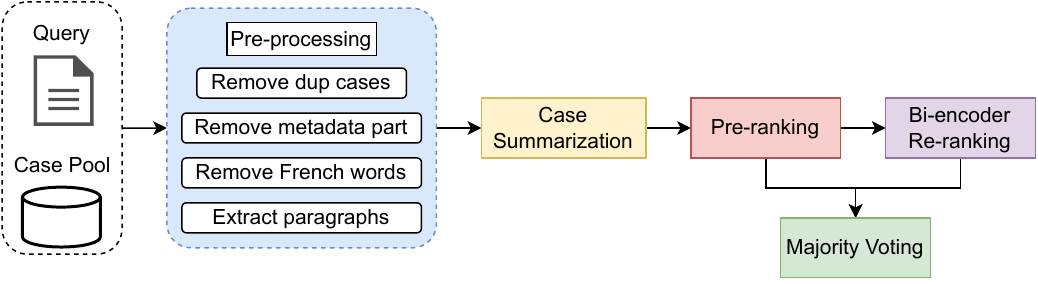}
    \caption{Overall architecture of multi-stage framework for Legal Case Retrieval.}
    \label{fig:task1_framework}
    \Description{task 1 framework}
\end{figure}

To overcome the challenges of Legal Case Retrieval, which are excessive length and logical structure of case documents, we proposed a four-stage framework involving LLMs for summarization and massive text embedding models for computing case relevance. The detailed architecture of the framework is presented in Figure \ref{fig:task1_framework}.

\paragraph{Data Pre-processing}
The case pool contained noise, including duplication and line break errors. Therefore, we performed a data processing step before training or computing case relevance. This process involved removing duplicate files and filtering out metadata, which included procedural details such as related parties and locations. Non-English sentences and passages were also removed using the Langdetect tool. Finally, paragraphs containing special placeholders were extracted. Through this pre-processing, we aimed to minimize irrelevant information in the case documents, ensuring that only content relevant to the judgment remained.


\paragraph{Case Summarization}
To enhance the representation of case law documents, we applied abstractive summarization using an LLM with zero-shot prompting. Specifically, Qwen-2.5 is instructed to generate a concise summary of each case in the predefined format shown in Listing \ref{lst:task1_sum}. This LLM is trivial for processing long context due to its context length of up to 131,072 tokens and generation of 8,192 tokens. By leveraging the in-context learning ability of LLMs, we compressed legal cases while preserving key facts and essential information for relevance comparison. 

\begin{lstlisting}[caption={Zero-shot prompting for structured-based case summarization.},captionpos=top,label={lst:task1_sum}]
Summarize the following Federal Court decision containing these parts:
Introduction, Facts, Relevant provisions, Analyses, and Court's conclusion
Legal case:
{INPUT_CASE}
Generated summary:
\end{lstlisting}


\paragraph{Pre-ranking Model}
The pre-ranking step serves as the first filter to select the most relevant case within a large number of candidates in the database. Therefore, a pre-ranking model should satisfy both performance and recall score, which measure the ability to find all the relevant cases. 
The relevance score between a query case $q$ and a candidate case $d$ is calculated as: 
\begin{equation}
    s_p(q,d) = Sim(\textbf{h}_q,\textbf{h}_d)
\end{equation}
where $\textbf{h}_q$ and $\textbf{h}_q$ are the representation of query $q$ and candidate $d$ from the pre-ranking model. Specifically, we employed a pre-trained model, BGE-m3, due to its strong generalization and context length of up to 8,192 tokens. Top-$k$ candidates with the highest relevance score are selected as the input for the re-ranking step. 

\paragraph{Re-ranking Model}
We employed two methods for the re-ranking phase: (1) fine-tuned a text embedding model--BGE-m3 on the COLIEE training set and (2) utilized a recent decoder-only model for text encoding--LLM2Vec. The BGE-m3 model is trained to distinguish relevant cases from irrelevant ones. Each retrieval method is expected to assign a higher score to a query’s positive samples compared to its negative ones. To achieve this, the training process minimizes the InfoNCE loss, defined as follows:
\begin{equation}
    \mathcal{L}_{InfoNCE} = -\log \frac{\exp(s(q,p^{+})/\tau)}{\sum_{p\in \{p^{+},p^{easy-},p^{hard-}\}}\exp(s(q,p)/\tau)}
\end{equation}
where $p^{+}$, $p^{easy-}$, and $p^{hard-}$ stand for positive, easy negative, and hard negative samples to the query $q$; $s(\cdot)$ is a similarity function (e.g., the dot product or cosine similarity), $\tau$ is the temperature coefficient. In this process, the positive samples are the ground truth labels, while the easy negative samples are selected randomly from the case pool. Hard negative cases are selected using a pre-ranking relevance score for effective model training.

Another re-ranking model is LLM2Vec, a decoder-only model for text encoding by performing three steps: enabling bi-directional attention, masked next token prediction, and unsupervised contrastive learning. The relevance score of a query $q$ and candidate $d$ is computed as:
\begin{equation}
    s_r(q,d)=Sim(\textbf{h}'_q,\textbf{h}'_d)
\end{equation}
where $\textbf{h}'_q$ and $\textbf{h}'_d$ are the representation of query $q$ and candidate $d$ from the re-ranking model, which could be fine-tuned BGE-m3 or LLM2Vec. 

\paragraph{Post-processing}
Finally, we combined the top-$m$ retrieved candidates from pre-ranking and re-ranking steps following the majority voting method to improve the performance and recall score. The voting operator can be defined as:
\begin{equation}
    Final\_case=mode(TopM(Pre),TopM(FtBGE),TopM(LLM))
\end{equation}
where $TopM(\cdot)$ is the function that returns top cases with the highest relevance score. 

\subsection{Experimental Setup}
The training set of COLIEE 2025 contains 7,350 cases, with an average of 4.1 relevant cases per query. The testing set consists of 2,159 files in the case pool and 400 cases for querying. The proposed method is evaluated on the COLIEE 2024 benchmark comprising 400 query cases and 1,734 documents in the pool. The official evaluation metrics for Legal Case Retrieval are precision, recall, and micro-F1. 

All models were implemented using Python and HuggingFace platform. Qwen-2.5-14B-Instruct\footnote{\url{https://huggingface.co/Qwen/Qwen2.5-14B-Instruct}} is utilized for case summarization due to its strengths in handling long context input. Massive text encoder model BGE-m3 was deployed for both pre-ranking and re-ranking phases. LLM2Vec-Meta-Llama-8B-Instruct\footnote{\url{McGill-NLP/LLM2Vec-Meta-Llama-3-8B-Instruct-mntp}} is another model used for the re-ranking phase. 
For the fine-tuning process, BGE-m3 was trained for 4 epochs, with a batch size of 8 and an initial learning rate of $1e^{-5}$. The number of hard and easy negative samples is 3. The majority voting step combines the top 10 cases from each model. 

Based on the evaluation results, we submitted three settings as follows:
\begin{itemize}
    \item Run 1: Top-5 candidates from re-ranking using fine-tuned BGE-m3.
    \item Run 2: Top-5 candidates from re-ranking using pre-trained LLM2Vec
    \item Run 3: Majority voting of pre-ranking, fine-tuned BGE-m3, and LLM2Vec outputs. 
\end{itemize}

\subsection{Result and Discussion}
Table \ref{tab:task1_prerank} presents the recall performance of the pre-ranking stage under two settings: with and without the case summarization step. Retrieval at top-200 and top-500 shows strong performance, successfully retrieving between 78\% and 89\% of the ground truth cases. The summarization step further improves recall across most metrics, though a slight drop is observed at $R@1$. Balancing retrieval effectiveness and computational efficiency, we select the top 200 returned cases from the pre-ranking stage as candidates for the subsequent retrieval process.

\begin{table*}[]
\caption{The recall performance of pre-ranking phases on the evaluation set.}
\label{tab:task1_prerank}
\begin{tabular}{lcccccccc}
\hline
            & \textbf{R@1 }   & \textbf{R@2}    & \textbf{R@5}    & \textbf{R@10}   & \textbf{R@100}  & \textbf{R@200}  & \textbf{R@500}  & \textbf{R@1000} \\ \hline
w/o summary & 0.0755 & 0.1133 & 0.1939 & 0.2797 & 0.6280 & 0.7387 & 0.8687 & 0.9507 \\
summary  & 0.0723 & 0.1165 & 0.2099 & 0.2970 & 0.6824 & 0.7784 & 0.8950 & 0.9654 \\ \hline
\end{tabular}
\end{table*}

Table \ref{tab:task1_eval} reports the performance of the proposed methods on the COLIEE 2024 benchmark. Among them, the ensemble run yields the best results across all metrics, with a particularly notable improvement in recall. Compared to the baseline, defined as the top-5 results from the pre-ranking phase, the proposed method improves the F1 score by 5\%, demonstrating the effectiveness of combining ranking and representation techniques. Notably, the pre-trained LLM2Vec model performs comparably to the fine-tuned BGE-m3, highlighting the potential of general-purpose language models in the legal case retrieval domain, even without task-specific fine-tuning. 

\begin{table}[ht]
\caption{Performance of the proposed methods on the evaluation set.}
\label{tab:task1_eval}
\begin{tabular}{lccc}
\hline
\textbf{Model}  & \textbf{F1}     & \textbf{Precision} & \textbf{Recall}  \\ \hline
Pre-ranking     & 0.1701          & 0.1515             & 0.1939           \\
BGE-m3-ft       & 0.2262          & 0.2015             & 0.2580           \\
LLM2Vec         & 0.2167          & 0.1930             & 0.2471           \\
Majority Voting & \textbf{0.2611} & \textbf{0.2153}    & \textbf{0.3316} \\ \hline
\end{tabular}
\end{table}

Table \ref{tab:task1_res} presents the official results of the Legal Case Retrieval task, which saw participation from seven teams with a total of 21 submitted runs. Our proposed framework, combining the strengths of a pre-ranking phase and fine-tuned text embedding models, secured fourth place on the leaderboard. As anticipated, the run using fine-tuned BGE-m3 (Run 1) outperformed the one based on the pre-trained LLM2Vec model (Run 2) by approximately 2\% across all metrics. The ensemble run, which integrates outputs from both approaches, achieved the highest performance among our submissions, demonstrating the benefit of combining complementary methods.
\begin{table}[ht]
\caption{Leaderboard of the Legal Case Retrieval task.}
\label{tab:task1_res}
\begin{tabular}{lccc}
\hline
\textbf{Team} & \textbf{F1}           & \textbf{Precision}    & \textbf{Recall}       \\ \hline
JNLP                     & 0.3353       & 0.3042       & 0.3735       \\
UQLegalAI                & 0.2962       & 0.2908       & 0.3019       \\
AIIR Lab                 & 0.2171       & 0.2040       & 0.2319       \\
{\ul NOWJ\_run3}         & {\ul 0.1984} & {\ul 0.1670} & {\ul 0.2445} \\
{\ul NOWJ\_run1}         & {\ul 0.1708} & {\ul 0.1605} & {\ul 0.1825} \\
{\ul NOWJ\_run2}         & {\ul 0.1580} & {\ul 0.1485} & {\ul 0.1688} \\
OVGU                     & 0.1498       & 0.1743       & 0.1313       \\
UB\_2025                 & 0.1363       & 0.1955       & 0.1046       \\
SIL                      & 0.0058       & 0.0054       & 0.0063       \\ \hline
\end{tabular}
\end{table}

\section{Task 2: Legal Case Entailment}
\subsection{Task Overview}
Given a decision $d$ and a relevant case $R=\{p_1,p_2,...,p_n\}$,  Task 2 aims to identify the specific paragraph $p\in R$ that entails the decision $d$. This task presents a fine-grained challenge in legal text understanding, as multiple paragraphs may reference related legal issues without directly supporting the decision. Unlike Task 1, which operates at the case level, Task 2 evaluates textual entailment at the paragraph level, using the same metrics.

The dominance of monoT5-based models among top-performing teams in COLIEE 2024 \cite{coliee_2024} underscores their effectiveness for legal entailment tasks. The AMHR team \cite{AMHR_2024} achieved the highest performance by fine-tuning a monoT5 model (pre-trained on MSMARCO), enhanced with hard negatives selected via BM25 and further refined using a score-ratio threshold tuned via grid search. Similarly, CAPTAIN \cite{CAPTAIN_2024} fine-tuned monoT5 with hard negative sampling, then selected top-$k$ candidate paragraphs to construct zero-shot and few-shot prompts for FlanT5-based in-context learning. The JNLP team \cite{JNLP_2024} also fine-tuned monoT5 on the task dataset and incorporated FlanT5 and Mixtral models for prompting-based inference.

\subsection{Methodology}

\begin{figure}
    \centering
    \includegraphics[width=1\linewidth]{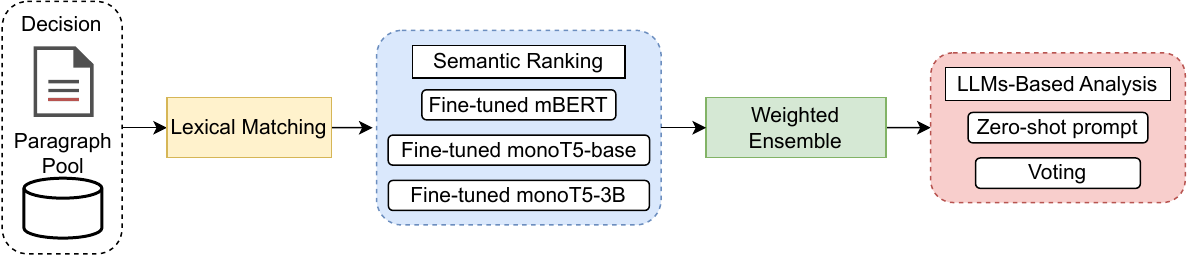}
    \caption{The three-stage framework based on lexical matching, language models, and LLMs for Legal Case Entailment.}
    \label{fig:task2_framework}
    \Description{task 2 framework}
\end{figure}

Our system follows a three-stage pipeline to identify the most relevant paragraphs that entail a given decision.

\textit{Lexical Pre-ranking:} 
In the first stage, we employ BM25 to retrieve paragraphs with high lexical overlap with the decision efficiently. This serves as a computationally inexpensive filtering step, which helps reduce the candidate pool while maintaining high recall of potentially entailing paragraphs.

\textit{Semantic Re-ranking:}
Next, we apply PLMs such as BERT and monoT5 to re-rank the paragraphs retrieved by BM25. These models assess each $(q,p_i)$ pair to capture deep semantic relationships beyond lexical matching. To adapt PLMs for legal text, both BERT and monoT5 are fine-tuned using the Cross-Entropy (CE) loss function, defined as:

\begin{equation}
\mathcal{L}_{CE} = - \sum_{i=1}^{N} \left[ y_i \log(\hat{y}_i) + (1 - y_i) \log(1 - \hat{y}_i) \right]
\end{equation}

During training, each query-paragraph pair $(q,p_i)$ is labeled as either positive relevant, $y_i = 1$ or negative non-relevant, $y_i = 0$. Negative labels are obtained by randomly sampling paragraphs from the set of paragraphs not relevant to the given query.  The models learn to differentiate these labels by minimizing the CE Loss, thus effectively enhancing their capability to distinguish between relevant and non-relevant paragraphs. After training, we combine the lexical and semantic scores to produce a refined ranking and select the top-$k$ paragraphs for further analysis. The value of $k$ is empirically determined to balance retrieval performance with computational efficiency.

\textit{LLM-Based Analysis:} 
In the final stage, we use LLMs to identify which paragraphs entail the decision. Rather than evaluating each ($q$, $p_i$) pair separately, we construct a single prompt that includes the decision $q$ and all $k$ candidate paragraphs with a clear instruction asking the model to determine which paragraphs support the decision. This holistic setup enables the LLM to consider inter-paragraph relationships and make more informed entailment judgments. The specific instruction is shown in Listing~\ref{lst:task2_prompt}, where we told the LLM to pick just one paragraph that best explains the decision's reasoning. They could only choose two paragraphs in cases where both were strictly necessary.

\begin{lstlisting}[caption={Zero-shot prompting for paragraph entailment.},captionpos=top,label={lst:task2_prompt}]
You are a legal expert tasked with identifying which paragraph(s) from a noticed case contain the reasoning or analysis that supports the decision of a new case. Below is the decision of the new case (query) and the paragraphs from the noticed case. Your task is to identify the single best paragraph that provides the reasoning or analysis leading to the decision. Only if there are two paragraphs that are both significantly important, equally critical, and absolutely necessary, you may return two. Otherwise, return only one. Do not select the paragraph that states the final decision or order. Instead, focus on the paragraph(s) that contain the reasoning or analysis that supports the decision.

Query (Decision of the New Case):
{query}

Paragraphs from the Noticed Case:
{paragraphs}

Which paragraph(s) contain the reasoning or analysis that supports the decision (entails the decision)?  
\end{lstlisting}

\subsection{Experimental Setup}

We fine-tune three Transformer-based language models: mBERT\footnote{\url{https://huggingface.co/google-bert/bert-base-multilingual-cased}}, monot5-base\footnote{\url{https://huggingface.co/castorini/monot5-base-msmarco-10k}} and monot5-3b\footnote{\url{https://huggingface.co/castorini/monot5-3b-msmarco-10k}}. During training, we randomly select 5 negative samples from the top 20 BM25 results to help the models learn more stably, rather than immediately using potentially confusing hard negatives. The fine-tuning configuration consists of 3 epochs, a learning rate of $1e^{-5}$, and a batch size of 8. The scores from these language models and BM25 are combined using weights optimized via grid search to produce the re-ranked list. Finally, we experiment with state-of-the-art LLMs such as {DeepSeek-V3}\footnote{ \url{https://huggingface.co/deepseek-ai/DeepSeek-V3}} and {QwQ-32B}\footnote{\url{https://huggingface.co/Qwen/QwQ-32B}}.

We submitted three settings as follows:

\begin{itemize}

    \item \textbf{Run 1}: Starting from the top 20 BM25-retrieved paragraphs, we re-rank them using the combined relevance scores. A threshold tuned on the validation set is applied to this list, and all paragraphs with scores above the threshold are predicted as entailing.
    \item \textbf{Run 2}: Instead of applying a threshold, this run feeds the 20 re-ranked paragraphs along with the decision query into {QwQ-32B} to get the final prediction. 
    \item \textbf{Run 3}: We expand the input to the top 35 candidates and prompt both {DeepSeek-V3} and {QwQ-32B} for entailment prediction. A voting strategy is then applied: only paragraphs selected by both models are kept. In cases of complete disagreement, both predictions are retained to ensure coverage.

\end{itemize}

\subsection{Result and Discussion}


Following Table \ref{tab:task2leaderboard}, the results suggest a progression in effectiveness based on the methodology employed. Initially, relying solely on combining reranking models based on relevance correlation appears to have limitations. These methods often evaluate passages individually, observing only a single paragraph in relation to the entailment during their core inference. This can lead to difficulties in distinguishing between paragraphs that are merely topically similar versus those that directly support or refute an entailment. Furthermore, such approaches may struggle when multiple distinct paragraphs are truly relevant, as they might overly focus on the single highest-scoring match rather than identifying a comprehensive set of evidence.

Following this initial phase, the introduction of an LLM for re-evaluating a curated list of potentially relevant passages (like the top 20 in Run 2) demonstrated a clear improvement, particularly in precision. By processing multiple candidate sentences together, the LLM can leverage broader context, leading to a more nuanced assessment of relevance and filtering out some spurious correlations identified by the initial re-rankers.

The most successful strategy involved both expanding the candidate pool (top 35) and implementing a stricter validation mechanism using two distinct LLMs in a voting agreement (Run 3). Expanding the pool likely increased the chances of capturing all necessary evidence while requiring consensus between two LLMs to act as a strong filter against false positives. This combination significantly boosted precision, indicating a higher confidence in the selected evidence and ultimately leading to the best overall F1 performance observed. This highlights the power of combining a wider search with multi-perspective LLM-based verification to enhance both the reliability and accuracy of the final results.
 
\begin{table}[ht]
\caption{Leaderboard of the Case Textual Entailment task.}
\label{tab:task2leaderboard}
\begin{tabular}{lccc}
\hline
\textbf{Team} & \textbf{F1}           & \textbf{Precision}    & \textbf{Recall} \\ \hline
{\ul NOWJ (run3)}                 & {\ul \textbf{0.3195}} & {\ul \textbf{0.3788}} & {\ul 0.2762}    \\
{\ul NOWJ\_(run2)}                & {\ul 0.2865}          & {\ul 0.2976}          & {\ul 0.2762}    \\
{\ul NOWJ\_(run1)}                & {\ul 0.2782}          & {\ul 0.2650}          & {\ul 0.2928}    \\
OVGU                              & 0.2454                & 0.2759                & 0.2210          \\
JNLP                              & 0.2412                & 0.2000                & \textbf{0.3039} \\
AIIR                              & 0.2368                & 0.2927                & 0.1989          \\
CAPTAIN                           & 0.1882                & 0.2547                & 0.1492          \\
UA                                & 0.1778                & 0.2090                & 0.1547          \\ \hline
\end{tabular}
\end{table}

\section{Task 3: Statute Law Retrieval}
\subsection{Task Overview}

For each input legal question Q, sourced from the Japanese Bar Examination, participating systems are required to automatically retrieve from the Japanese Civil Code the complete set of articles  $\{a_1, a_2,\dots, a_n\}$ deemed relevant. Relevance is defined by the condition that an article, either individually or in combination with others, entails a Yes/No answer to question Q. Both the legal questions and the Civil Code corpus are provided in Japanese, accompanied by English translations.

Both the JNLP \cite{JNLP_2024} and CAPTAIN \cite{CAPTAIN_2024} utilized BERT-base-Japanese models fine-tuned for employing ensemble strategies across their submissions. JNLP initially generated a ranked list by ensembling predictions from multiple BERT checkpoints. Their subsequent runs involved distinct LLMs2 for post-processing: one run used Mistral with prompting for final selection, another employed RankLLaMA to re-score top candidate pairs, and the third utilized Orca and Qwen for list refinement, also incorporating results from the Mistral run to improve recall. Similarly, the CAPTAIN approach involved ensembling, initially using top BERT checkpoints. However, their refinement primarily focused on filtering results using LLM prompting with Flan T5, applied either to the BERT ensemble output or to results generated by a fine-tuned MonoT5 re-ranker. CAPTAIN's final submissions often resulted from ensembling these differently filtered result sets.

\subsection{Methodology}

\begin{figure}
    \centering
    \includegraphics[width=1\linewidth]{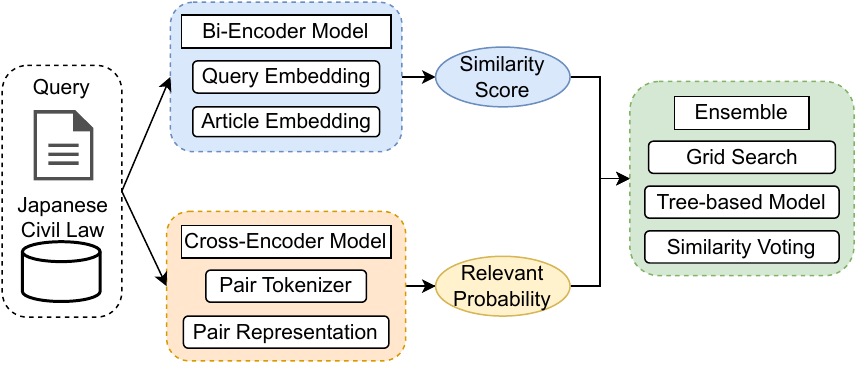}
    \caption{Overall architecture of the proposed framework for Statute Law Retrieval.}
    \label{fig:task3_framework}
    \Description{task 3 framework}
\end{figure}

Given the remarkable success and demonstrated semantic proficiency of contemporary open-source LLMs on diverse information retrieval and question-answering benchmarks, we posit that strategically combining these powerful pre-trained models presents a highly promising approach for addressing the specific demands of this legal retrieval task. Consequently, our methodology prioritizes the effective utilization and ensemble of existing publicly available LLM architectures instead of creating specialized models from scratch. The illustration of our proposed method is presented in Figure \ref{fig:task3_framework}.

\textit{Core Retrieval Architecture}: Our approach utilizes two main categories of models: Bi-Encoders and Cross-Encoders. For the Bi-Encoder group, models including bge, e5, stella, and NV-Embed are employed to generate independent vector representations (embeddings) for questions and legal articles separately. Subsequently, cosine similarity is calculated between these respective embedding pairs to produce an initial relevance ranking score. The Cross-Encoder group, using models like bge-reranker and gte-reranker variants, processes question-article pairs jointly, inputting them simultaneously into the model to directly compute a score reflecting their correlation or relevance.

\textit{Tree-based Ensemble}: Trains a LightGBM (LGBM) model using relevance scores from multiple base retrieval models as input features. The trained LGBM then predicts a final relevance score for each query-article pair, capturing non-linear relationships between the base model outputs.

\textit{Grid Search Weighted}: Identifies the top N performing base models on a development set. Performs a grid search to find optimal weights $(w_1, ..., w_N)$ that maximize performance when linearly combining the scores of these top models.
\begin{equation}
Score_{final} = \sum_{i=1}^{N}{w_i * Score_{Mi}} 
\end{equation}

\textit{Similarity-Informed Voting Ensemble}: For a given test query Q, first retrieves the most similar queries $Q_{sim}$ from historical data. Then, evaluates the performance of each base model $(M_i)$ on $Q_{sim}$. Finally, aggregates the predictions of the base models for Q, weighting each $M_i$ contribution based on its measured historical accuracy on similar queries $Q_{sim}$.
    
\subsection{Experimental Setup}


Our approach utilizes a diverse set of pre-trained Bi-Encoder and Cross-Encoder models from the Hugging Face Hub. Bi-Encoders are used for efficient initial candidate retrieval, while Cross-Encoders refine these results through more precise re-ranking. To enhance overall performance, we employ three ensemble strategies that combine the relevance scores produced by these models for our final submissions.













We submitted three runs based on different ensemble techniques:
\begin{itemize}

    \item \textbf{Run 1:} Employed a LightGBM model trained on scores derived from \textit{all} base models to predict final relevance.
    \item \textbf{Run 2:} Applied optimized linear weights, found via grid search, to combine scores from three selected top-performing base models: \textit{NV-Embed-v1, multilingual-e5-large-instruct, bge-reranker-large}.
    \item \textbf{Run 3:} Used query similarity, calculated via bi-encoder model, \textit{multilingual-e5-large}, to dynamically weight base model predictions based on historical performance on similar queries.
\end{itemize}

\subsection{Result and Discussion}

\begin{table}[]
\caption{Results of Statute Law Retrieval task in COLIEE 2025.}
\label{tab:task3result}
\begin{tabular}{lccc}
\hline
  \textbf{Team}                      & \textbf{F2}     & \textbf{Precision} & \textbf{Recall} \\ \hline
JNLP\_RUN1              & \textbf{0.8365} & 0.8037	    & 0.8744 \\
CAPTAIN.H2              & 0.8301 & \textbf{0.8333}    & 0.8516 \\
INFA                    & 0.6917 & 0.7671    & 0.6826 \\
mpnetAIIRLab            & 0.6674 & 0.3562    & \textbf{0.8858} \\
OVGU3                   & 0.6041 & 0.6347	    & 0.6142 \\
UIwa                    & 0.5816 & 0.5856    & 0.5890 \\
UA-mpnet                  & 0.2540 & 0.0986	   & 	0.4361 \\ \hline
\textbf{Our runs }               &        &           &        \\
NOWJ.H2                 & 0.7702 & 0.7572    & 0.8086 \\
NOWJ.H1                 & 0.7311 & 0.7352    & 0.7511 \\
NOWJ.H3                 & 0.7069 & 0.7534    & 0.7100 \\ \hline
\end{tabular}%
\end{table}

The results of our ensemble method are presented in the Table \ref{tab:task3result}. Analysis of the results reveals that the Run 2 method achieved the highest efficacy F2, which is 0.7702, leveraging a strategy centered on the linear combination of scores from only three top-performing base models, with weights optimized via grid search. This suggests that the selection of high-quality input signals coupled with meticulous optimization, even when employing a straightforward methodology, proved pivotal for success, particularly with the high Recall score of 0.8086, making a substantial contribution to the F2 metric. Run 1, which employed a more sophisticated LightGBM model across all base models, yielded suboptimal performance, F2 is 0.7311, potentially attributable to the introduction of noise from less effective models or inherent challenges in optimizing this potent ensemble model over the larger set of inputs. Run 3 proposed a more advanced concept involving dynamic weight adjustment predicated on query similarity, yet exhibited the lowest performance, F2, which is 0.7069. The inherent complexity of this approach, compounded by potential difficulties in precisely quantifying query similarity or in effectively leveraging historical performance data, seemingly impeded its practical efficacy relative to the more straightforward ensemble strategies.

\section{Task 4: Legal Textual Entailment}

\subsection{Task Overview}
This task aims to develop a Yes/No question-answering system given a legal question $q$ and relevant articles $A=\{a_1,...,a_n\}$, in which $n \geq 1$. The training set contains triplets $\{q,A,label\}$, in which $label\in \mathbb{R}^2,label=\{Y,N\}$. In the inference phase, the system is assessed by answering unseen queries. The official evaluation metric for this task is Accuracy, which is computed by the number of correct answer queries divided by the total number of queries. 

Last year, most teams utilized the in-context learning capabilities of various LLMs (e.g., FlanT5, Qwen, Llama, GPT-3.5) and combined their outputs for post-processing. The CAPTAIN team\cite{CAPTAIN_2024} secured first place with data augmentation and LLM fine-tuning using LoRA. The runner-up team, JNLP\cite{JNLP_2024}, experimented with prompting across multiple LLMs—Qwen 14B, Mistral 7B, Flan-Alpaca, and FlanT5—employing a voting ensemble approach. 
Therefore, to prevent data leakage, i.e., test queries appearing in the model's training set, only open-source LLMs released before July 2024 are allowed in this year's competition.

\subsection{Methodology}
\begin{figure}[ht]
    \centering
    \includegraphics[width=1\linewidth]{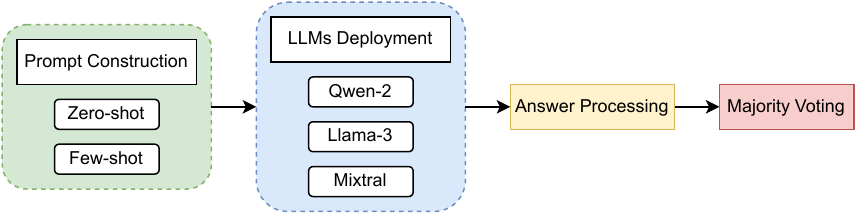}
    \caption{The overall LLMs-based framework for Legal Textual Entailment.}
    \label{fig:task4_framework}
    \Description{task 4 framework}
\end{figure}

Inspired by recent advancements in legal text processing \cite{coliee_2024,10.1007/978-981-97-3076-6_13,10.1145/3594536.3595176}, the proposed framework leverages the in-context learning capability of open-source LLMs (i.e. Qwen-2, Llama-3, Mixtral) to address the problem of textual inference in legal text. The framework contains four main phases: prompt construction, LLMs deployment, answer processing, and majority voting as presented in Figure \ref{fig:task4_framework}.

\paragraph{Prompt Construction}: We construct the prompt collection following both zero-shot and few-shot, parallel prompting. For few-shot prompting, article-shared questions are selected as examples for the LLMs. If the query shares no article with others, the examples are identified based on semantic similarity, computed by a bi-encoder architecture. 


Finally, the LLM input can be defined as follows:
\begin{equation}
    Input = [System, Inst, Premise, Hypothesis, Examples]
\end{equation}
in which $System$ is the system prompt (e.g ``You are an overthinking legal assistant who always gives the best advice.''), $Inst$ is the step-by-step guidance to solve the task, $Examples$ are similar samples extracted from the training set, $Premise$ and $Hypothesis$ are replaced by relevant articles and question respectively.

\paragraph{LLMs Deployment}
Qwen-2, Llama-3, and Mixtral are chosen as the backbone models since they are the best LLMs that meet the release date constraint. In the prompt template, the placeholders {premise} and {hypothesis} are replaced with the question and article content. If multiple relevant articles exist, they are concatenated to form a single hypothesis base. Finally, LLM follows instructions and examples in the input to generate responses, which may include simple binary answers or explanations.

\paragraph{Answer Processing}
A scanning function is designed to extract binary answers based on specific patterns (e.g., ``TRUE'', ``FALSE''). Since responses include explanations, reasoning paths, and noise, the function first identifies the ``CONCLUSION'' section in the text before extracting the answer. If the extracted answer matches positive patterns, it returns ``Y''; otherwise, it returns ``N''.

\paragraph{Majority Voting}
Finally, the extracted answers are combined following the majority voting method to improve performance and reliability. The voting operator can be defined as follows:
\begin{equation}
    Final\_answer = mode(Ans_{Llama},Ans_{Qwen},Ans_{Mix})
\end{equation}

\subsection{Experimental Setup}
The COLIEE 2025 benchmark includes 1,206 training samples and 74 testing samples, extracted from the Japanese Bar Exam. The proposed method is evaluated on the COLIEE 2023 and 2024 benchmarks containing 101 and 109 samples, respectively. The official evaluation metric for Legal Textual Entailment is accuracy.

We implemented the 4-bit quantized version of Qwen-2.5-72B-Instruct\footnote{\url{https://huggingface.co/Qwen/Qwen2-72B-Instruct}}, Llama-3-70B-Instruct\footnote{\url{meta-llama/Meta-Llama-3-70B-Instruct}}, and Mixtral-8x7B-Instruct-v0.1\footnote{\url{mistralai/Mixtral-8x7B-Instruct-v0.1}}. The temperature is set at 0 to ensure consistency among generations. The maximum length of responses is 800 tokens. Based on the evaluation results, we prepare three settings for submissions as follows:
\begin{itemize}
    \item Run 1: Qwen-2 and legal few-shot prompting.
    \item Run 2: Majority voting of Qwen-2, Llama-3, Mixtral and legal zero-shot prompting.
    \item Run 3: Majority voting of Qwen-2, Llama-3, Mixtral and legal few-shot prompting.
\end{itemize}

\subsection{Result and Discussion}
Table \ref{tab:task4_eval} presents the evaluation results of the proposed method on the development set. The highest accuracy is achieved using legal few-shot prompting, scoring 0.8217 on COLIEE 2023 and 0.8623 on COLIEE 2024. However, the effectiveness of few-shot prompting is inconsistent and unreliable. Additionally, LLM performance remains similar across most experiments. To improve robustness, we combined LLM outputs using the majority voting method and submitted both few-shot and zero-shot prompting runs.

\begin{table}[ht]
\caption{Results of the proposed method on the evaluation set.}
\label{tab:task4_eval}
\begin{tabular}{lccc}
\hline
\textbf{Model}                 & \multicolumn{1}{l}{\textbf{Few-shot}} & \multicolumn{1}{l}{\textbf{COLIEE 2023}} & \multicolumn{1}{l}{\textbf{COLIEE 2024}} \\ \hline
Qwen-2-72B   & \ding{51}                            & 0.7227                          & \textbf{0.8623}                          \\
   &   \ding{55}                           & 0.7326                          & 0.8073                          \\
Llama-3-70B  & \ding{51}                             & 0.7326                          & 0.7522                          \\
  &  \ding{55}                           & 0.7920                          & 0.7889                          \\
Mixtral-8x7B & \ding{51}                             & \textbf{0.8217 }                         &     0.7339                            \\
 &  \ding{55}                           & 0.7722                          & 0.7981                          \\ \hline
\end{tabular}
\end{table}

The official ranking for the Legal Textual Entailment task is shown in Table \ref{tab:task4_res}. This year, 10 teams participated, submitting a total of 30 runs. Unexpectedly, all of our runs achieved an accuracy score of 0.7397, securing 6th place on the leaderboard. Despite their strong in-context learning abilities, LLMs still struggle with real-world challenges such as Legal Textual Entailment in COLIEE. Future work should focus on enhancing the performance and robustness of LLM-based methods through data augmentation and fine-tuning, particularly in handling complex legal reasoning and reducing inconsistencies in model predictions.

\begin{table}[t]
\caption{The official leaderboard of the Legal Textual Entailment task.}
\label{tab:task4_res}
\begin{tabular}{lcc}
\hline
\textbf{Team} & \textbf{Correct} & \textbf{Accuracy} \\ \hline
KIS3          & 66               & 0.9041            \\ 
CAPTAIN2      & 60               & 0.8219            \\ 
JNLP002       & 59               & 0.8082            \\ 
UA2           & 57               & 0.7808            \\ 
KLAP.H2       & 56               & 0.7671            \\ 
{\ul NOWJ.run1} & {\ul 54}         & {\ul 0.7397}      \\ 
{\ul NOWJ.run2} & {\ul 54}         & {\ul 0.7397}      \\ 
{\ul NOWJ.run3} & {\ul 54}         & {\ul 0.7397}    \\
OVGU1         & 54               & 0.7397           \\ 
RUG\_V1       & 48               & 0.6575            \\ 
AIIRLLaMA     & 44               & 0.6027            \\ 
BaseLine      & 37               & 0.5068            \\ \hline
\end{tabular}
\end{table}

\section{Pilot Task: Legal Judgment Prediction}
\subsection{Task Overview}
This task addresses legal judgment prediction in Japanese tort cases, where plaintiffs claim that defendants' actions constitute a tort, while defendants contest these claims. It consists of two sub-tasks:  
\begin{itemize}
\item \textbf{Tort Prediction (TP)}: Given undisputed facts ($U$) and arguments from plaintiffs ($P$) and defendants ($D$), the goal is to predict whether the judge affirms the tort ($T$, a Boolean value). {TP} is evaluated using \textit{accuracy}, measuring the proportion of correctly predicted cases.  
\item \textbf{Rationale Extraction (RE)}: Identifies which arguments were accepted by the judge. The task predicts Boolean sequences ($R^P, R^D$), indicating accepted arguments in $P$ and $D$. {RE} is evaluated using the \textit{micro-F1 measure}.  
\end{itemize}  
In short, \textit{Pilot Task} requires predicting $(T, R^P, R^D)$ from $(U, P, D)$.

\subsection{Methodology}
A brief data analysis was conducted to understand the dataset. 
In this task, a training set and a test set are provided by the organizers. The dataset is provided in JSONL format, with each case consisting of undisputed facts, claims from both parties and the final court decision. In the test set, the accepted status of claims and the final court decision are hidden and require predictions.

\begin{table}[]
\caption{Statistics of data in the Legal Judgment Prediction task.}
\label{tab:Pilot Task analysis}
\centering
\begin{tabular}{@{}lcc@{}}
\toprule
\multicolumn{1}{c}{}                          & \textbf{Train set} & \textbf{Test set} \\ \midrule
No. samples                                   & 6508      & 812      \\
Average facts / Case                          & 1.33      & 1.37     \\
Max facts / Case                              & 134       & 26       \\
Average plaintiff claims / Case               & 3.87      & 3.82     \\
Max plaintiff claims / Case                   & 111       & 130      \\
Average defendant claims / Case               & 3.45      & 3.49     \\
Max defendant claims / Case                   & 86        & 50       \\
No. samples without facts                     & 3764      & 414      \\
No. samples without plaintiff claims          & 454       & 33       \\
No. samples without defendant claims          & 1321      & 106      \\
No. samples without both claims               & 144       & 1        \\
No. samples without facts and both claims     & 1         & 0        \\
No. samples with facts and both claims        & 2215      & 335      \\ \bottomrule
\end{tabular}
\end{table}

As shown in Table \ref{tab:Pilot Task analysis}, the training set is more detailed, containing a notably higher maximum number of facts per case, with 134 facts, compared to the test set, which only has 26. Additionally, the training set includes a larger proportion of cases with missing attributes, with 3764 cases missing undisputed facts, representing approximately 58\% of the training dataset. Only 2215 cases, or 34\% of the training dataset, are complete with all three attributes, which may impact the model's ability to generalize.

\begin{figure}
    \centering
    \includegraphics[width=1\linewidth]{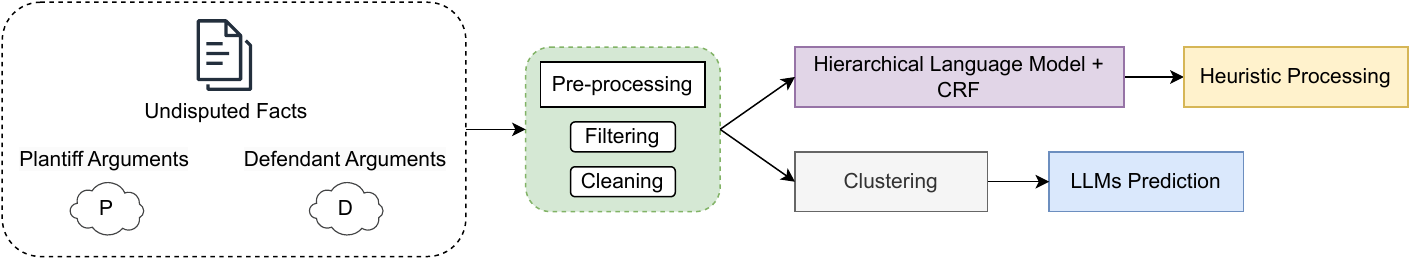}
    \caption{The overall architecture for Legal Judgment Prediction.}
    \label{fig:task5_framework}
    \Description{task 5 framework}
\end{figure}

Figure \ref{fig:task5_framework} illustrates the overall architecture of our proposed framework for Legal Judgment Prediction. We adopt two different approaches. The first utilizes a hierarchical language model combined with a Conditional Random Field (CRF) layer, followed by a heuristic-based post-processing step designed to refine predictions. The second approach leverages the reasoning capabilities of advanced large language models (LLMs), which are prompted to perform judgment prediction on clustered case inputs.
.

\textit{Pre-processing:} The dataset contains samples that may be missing one, two, or all three attributes: undisputed facts, plaintiff claims, or defendant claims. In the training set, we remove samples that are missing two or more of these attributes and retain the rest. This results in 1246 samples being discarded due to missing two or more attributes, which accounts for about 19\% of the original training set. The remaining samples are used for further processing using our methods.

Our primary approach employs a hierarchical language model, following the Inter-Span Transformer (IST) architecture \cite{Yamada_2024}, which captures both word-level and span-level representations of legal texts. At the word-level encoder, we use {ModernBERT-Ja-310M} \footnote{\url{https://huggingface.co/sbintuitions/modernbert-ja-310m}}, a large variant of ModernBERT \cite{modernbert_2024} trained on a high-quality corpus of Japanese and English texts. ModernBERT \cite{modernbert_2024}, an improved version of BERT\cite{bert_2019}, integrates local and global attention mechanisms to efficiently handle long sequences while maintaining computational efficiency. It also incorporates Rotary Positional Embeddings (RoPE) \cite{rope_2024}, further improving performance across various NLP tasks. Claims and facts of each tort are first processed through the word-level encoder to generate contextual embeddings. The span-level Transformer \cite{transformers_2017} then captures interactions among claims, where each claim representation is enriched with fact, party-type, and positional embeddings. The TP task is framed as a binary classification problem, while RE is treated as a sequence labeling task. To model dependencies between closely related claims, particularly those from the same side, we incorporate a Conditional Random Field (CRF) layer \cite{crf_2001}, ensuring that claim predictions remain consistent within a party's argument.

To refine predictions, we introduce post-processing heuristics for both sub-tasks. In TP, if one party has both more accepted and fewer unaccepted claims than the other, the predicted decision is reversed to favor the dominant side. Once the final TP decision is established, RE predictions are refined by ensuring consistency within each party: if the number of accepted claims exceeds unaccepted ones by at least $x$ times (optimized via grid search), the unaccepted claims are adjusted to accepted, and vice versa.

To promote explainability, we propose a clustering-based approach that organizes claims into semantically coherent subarguments. First, claims from both sides are embedded using the {paraphrase-multilingual-MiniLM-L12-v2} model\footnote{\url{https://huggingface.co/sentence-transformers/paraphrase-multilingual-MiniLM-L12-v2}}, a pretrained sentence transformer \cite{sbert_2019}. The embeddings are then clustered using HDBSCAN \cite{hdbscan_2013}, a density-based algorithm that groups similar claims while leaving outliers unclustered. Undisputed facts are incorporated into these clusters to provide additional context. Each subargument is independently assessed by {DeepSeek-V3}\footnote{\url{https://huggingface.co/deepseek-ai/DeepSeek-V3}}, a large language model, to generate predictions for both TP and RE. The final TP decision of the tort is determined through a voting mechanism, where the side with the most winning subarguments prevails. For RE, unclustered claims inherit the majority stance of their respective party to maintain consistency. The entire case is treated as a single cluster if no clusters are formed.

\subsection{Experimental Setup}
Our primary approach employed a hierarchical transformer architecture using ModernBERT-Ja-310M as the word-level encoder, followed by a span-level Transformer with 8 layers and 8 attention heads. The model was trained for 18 epochs using the AdamW\cite{adamw_2019} optimizer with a learning rate of $6e^{-6}$ and a linear warmup schedule over 10\% of the total training steps. We combined binary cross-entropy loss for TP and CRF loss for RE, weighted by a factor $\alpha=0.4$. Each input case was processed with a maximum of 64 claims, where individual claims were truncated to 64 tokens and the aggregated undisputed facts were limited to 512 tokens. The decision threshold for TP classification was optimized via grid search on the validation set with an optimal value of 0.3838. The RE post-processing threshold was also determined through grid search, with $x=2$ selected as the optimal value. For our clustering-based approach, we generated claim embeddings using \textit{paraphrase-multilingual-MiniLM-L12-v2} and performed HDBSCAN clustering with at least two claims per cluster, using cosine similarity as the distance metric.

We submitted three settings as follows:

\begin{itemize}
\item \textbf{Run 1}: We utilize the hierarchical language model architecture with ModernBERT-Ja-310M and a span-level Transformer, predicting TP as binary classification and RE with a CRF layer, without additional post-processing.
\item \textbf{Run 2}: We enhance Run 1 by incorporating the outlined post-processing heuristics.
\item \textbf{Run 3}: We implement the proposed clustering-based method, embedding claims with \textit{paraphrase-multilingual-MiniLM-L12-v2}, clustering via HDBSCAN, and assessing clusters with DeepSeek-V3 for TP and RE predictions, followed by our proposed voting mechanism to determine the final TP decision.
\end{itemize}

\subsection{Result and Discussion}

\begin{table}[t]
\caption{The official leaderboard of the Legal Judgment Prediction task.}
\centering
\label{tab:pilot_results}
\begin{subtable}[t]{0.22\textwidth}
\centering
\begin{tabular}{lc}
\hline
\textbf{Team}      & \textbf{Accuracy} \\ \hline
CAPTAIN            & \textbf{76.5\%}   \\
KIS                & 71.3\%            \\
{\ul NOWJ (run 2)} & {\ul 67.1\%}      \\
omega              & 66.6\%            \\
{\ul NOWJ (run 1)} & {\ul 63.8\%}      \\
{\ul NOWJ (run 3)} & {\ul 59.7\%}      \\
OVGU               & 55.3\%            \\
LLNTU              & 54.1\%            \\ \hline
\end{tabular}
\subcaption{Tort Prediction}
\label{tab:tp}
\end{subtable}%
\hfill
\begin{subtable}[t]{0.22\textwidth}
\centering
\begin{tabular}{lc}
\hline
\textbf{Team}      & \textbf{F1 score} \\ \hline
KIS                & \textbf{71.2\%}   \\
CAPTAIN            & 70.6\%            \\
{\ul NOWJ (run 2)} & {\ul 69.2\%}      \\
omega              & 69.1\%            \\
LLNTU              & 68.2\%            \\
{\ul NOWJ (run 1)} & {\ul 68.1\%}      \\
OVGU               & 65.7\%            \\
{\ul NOWJ (run 3)} & {\ul 55.9\%}      \\ \hline
\end{tabular}
\subcaption{Rationale Extraction}
\label{tab:re}
\end{subtable}
\end{table}

Our heuristic post-processing step in run 2 yielded notable improvements, achieving 67.1\% accuracy in TP and 69.2\% F1 score in RE, representing gains of 3.3 and 1.1 percentage points, respectively, over run 1. These improvements validate our hypothesis that maintaining consistency between claim patterns and final decisions can enhance model performance. Our clustering-based approach (run 3) underperformed with 59.7\% TP accuracy and 55.9\% RE F1 score, suggesting that while semantic clustering provides explainability, it may oversimplify the complex relationships in legal argumentation. This performance gap highlights the importance of structural dependencies in legal reasoning. Consequently, our best approach (run 2) secured third place in both sub-tasks, with a 9.4 percentage point gap in TP accuracy and a 2.0 percentage point gap in RE F1 score compared to the top-performing team. Future work could consider alternative model architectures, integrating legal-reasoning prompts with LLMs, and developing structured clustering methods to enhance both performance and explainability.

\section{Conclusion}
In COLIEE 2025, the NOWJ team successfully developed and deployed innovative methodologies across all five competition tasks, notably securing the highest performance in Task 2 (Legal Case Entailment) by integrating hierarchical retrieval methods with contextualized re-ranking using Large Language Models (QwQ-32B and DeepSeek-V3). Our multi-stage ensemble framework, combining embedding-based models and advanced LLM-based techniques, consistently showed effectiveness in managing complex legal reasoning and retrieval scenarios. Results demonstrate the clear advantage of combining embedding precision, transformer-based summarization, and deep contextual reasoning through generative language models. Future research should continue exploring model fine-tuning, improved ensemble strategies, and enhanced prompt engineering to further advance the interpretability, accuracy, and generalizability of legal information systems.


\bibliographystyle{ACM-Reference-Format}
\bibliography{sample-base}

\appendix

\end{document}